\documentclass[runningheads]{llncs}
\usepackage[T1]{fontenc}
\usepackage{graphicx}
\usepackage{hyperref}
\usepackage{color}

\usepackage{pifont}
\usepackage[table,xcdraw]{xcolor}
\usepackage{caption}
\usepackage{amsmath}
\usepackage{amsbsy}
\usepackage{amsfonts}
\usepackage{amssymb}
\usepackage{array}
\usepackage[misc]{ifsym}
\usepackage{subcaption}
\usepackage{float}

\begin{document}
\title{Spatiotemporal Representation Learning for Short and Long Medical Image Time Series}
\titlerunning{Spatiotemporal Representation Learning}
%

\author{Chengzhi Shen \inst{1}$^\text{(\Letter)}$, 
        Martin J. Menten \inst{1,7}, 
        Hrvoje Bogunović \inst{2}, 
        Ursula Schmidt-Erfurth \inst{2}, 
        Hendrik P.N. Scholl \inst{3,4,5}, 
        Sobha Sivaprasad \inst{6}, 
        Andrew Lotery \inst{7}, 
        Daniel Rueckert \inst{1,8}, 
        Paul Hager\thanks{Paul Hager and Robbie Holland contributed equally.} \inst{1}, 
        Robbie Holland$^*$ \inst{8}}
        
\authorrunning{C. Shen et al.}
%
\institute{Technical University of Munich, Germany \\\email{chengzhi.shen@tum.de} \and 
Laboratory for Ophthalmic Image Analysis, Medical University of Vienna, Austria \and 
Institute of Molecular and Clinical Ophthalmology Basel, Switzerland \and
Department of Ophthalmology, University of Basel, Switzerland \and
Department of Clinical Pharmacology, Medical University of Vienna, Austria \and
Moorfields Eye Hospital NHS Foundation Trust, London, United Kingdom \and
Clinical and Experimental Sciences, University of Southampton, United Kingdom \and
BioMedIA, Imperial College London, United Kingdom 
}
\maketitle              
\begin{abstract}
Analyzing temporal developments is crucial for the accurate prognosis of many medical conditions. Temporal changes that occur over short time scales are key to assessing the health of physiological functions, such as the cardiac cycle. Moreover, tracking longer term developments that occur over months or years in evolving processes, such as age-related macular degeneration (AMD), is essential for accurate prognosis. Despite the importance of both short and long term analysis to clinical decision making, they remain understudied in medical deep learning. State of the art methods for spatiotemporal representation learning, developed for short natural videos, prioritize the detection of temporal constants rather than temporal developments. Moreover, they do not account for varying time intervals between acquisitions, which are essential for contextualizing observed changes. To address these issues, we propose two approaches. First, we combine clip-level contrastive learning with a novel temporal embedding to adapt to irregular time series. Second, we propose masking and predicting latent frame representations of the temporal sequence. Our two approaches outperform all prior methods on temporally-dependent tasks including cardiac output estimation and three prognostic AMD tasks. Overall, this enables the automated analysis of temporal patterns which are typically overlooked in applications of deep learning to medicine. Code is available at \url{https://github.com/Leooo-Shen/tvrl}.

\end{abstract}
\section{Introduction}
Analyzing temporal developments in medical images is crucial to the practice of medicine. Clinicians analyze temporal change over \textit{short} time scales, such as arrhythmia in the cardiac cycle in MR videos \cite{bernard2018deep,komatsu2021detection,mondejar2019heartbeat} or prenatal movement in ultrasound \cite{pugash2008prenatal}, in order to analyze dynamic physiological processes and function. Clinicians also track the development of processes that evolve gradually, such as disease progression \cite{koch2013fear,jensen2014temporal,holland2023clustering} or a patient's response to treatment \cite{chainani2001oral,graham2014current}, over \textit{long} time frames. In this longitudinal data, images are acquired at sparse and typically irregular intervals that may be months or even years apart. Despite the importance of both temporal dynamics and historical observations to the present diagnosis and prognosis of patients, the development of representation learning methods capable of learning these trends in medical data has thus far been overlooked.

Current state of the art approaches for spatiotemporal learning \cite{feichtenhofer2021large,qian2021spatiotemporal} were primarily developed on the Kinetics dataset \cite{kay2017kinetics}, a human action recognition dataset which depicts only a single action or event per video. To predict this singular event, these methods prioritize the detection of video-level features that persist over time, and in doing so explicitly ignore changes between frames. As such, these methods are unsuitable for learning temporally variant features over the timeline. Furthermore, these methods were designed for videos with a fixed framerate and are thus unable to handle varying time intervals in irregularly sampled longitudinal sequences.

To address these issues, we contribute the following: 
\begin{itemize} 
\item We demonstrate that, while established strategies for representation learning in natural videos perform adequately on short cardiac videos, they are especially poor at modeling long-range developments in longitudinal retinal images due to the neglecting of temporally variant features. 
\item We then propose a simple clip-level contrastive learning strategy that uses a single clip with time embedding to encode varying time intervals, outperforming all natural video baselines.
\item Finally, we propose a novel method that combines the efficacy of clip-level contrastive learning with a frame-level latent feature prediction task. Our method outperforms all natural video baselines in cardiac output estimation and three prognostic tasks for late stage AMD.
\end{itemize}

\begin{figure}[t!]
    \centering
    \includegraphics[width=\linewidth]{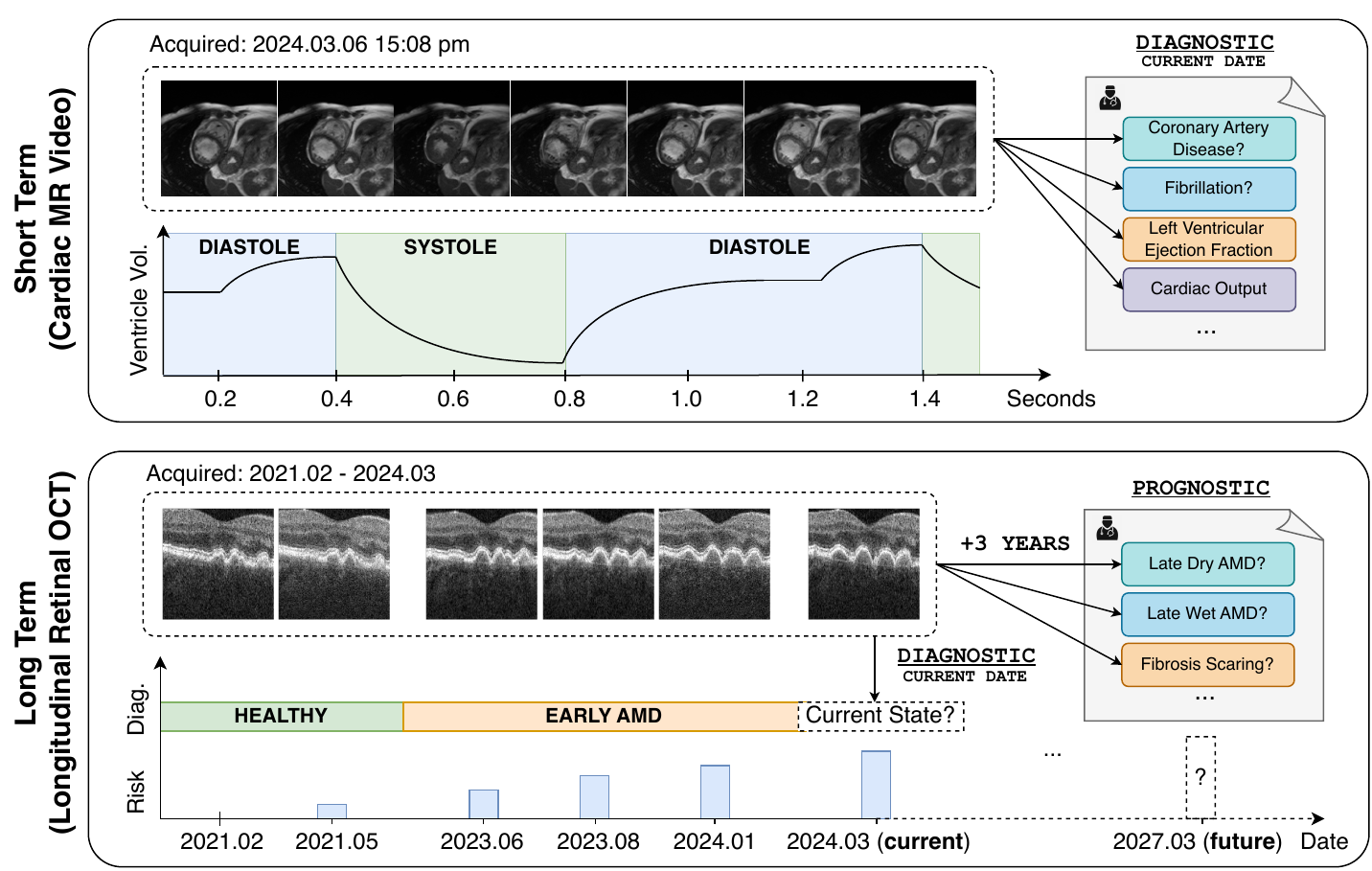}
    \caption{Clinicians use spatiotemporal data to observe temporal variations over both \textit{short} and \textit{long} time scales. To observe dynamic physiological processes such as a beating heart, short cardiac MR videos can be captured (top). Tracking long term developments, such as disease progression in retinal OCT scans, require longitudinal acquisitions that typically occur at irregular intervals of years (bottom). Modeling and extrapolating the trajectory of historical change is crucial for the prognosis of late stage disease.}
    \label{fig:data}
\end{figure}
\section{Related Work}
\subsection{Self-Supervised Video Representation Learning}
In the image domain, contrastive learning has emerged as a strong and robust method for learning global imaging features without labels \cite{chen2020simple,grill2020bootstrap,he2020momentum,chen2021exploring,caron2021emerging}. Similarly, contrastive approaches have achieved the state-of-the-art on the standard benchmark dataset for natural videos, Kinetics \cite{kay2017kinetics}. By minimizing the feature distance between differing segments, or clips, sampled from different time points in the same video, these approaches learn to extract the single human action recorded over the span of each video \cite{qian2021spatiotemporal,feichtenhofer2021large,wang2022long}. Specifically, Qian et al. \cite{qian2021spatiotemporal} adopt a probabilistic function to vary the time between the two sampled clips, Feichtenhofer et al. \cite{feichtenhofer2021large} sample up to four clips, and Wang et al. sample clips of varying lengths \cite{wang2022long}. However, by learning features that persist across the entire video, these methods ignore temporal variations which are associated with many medical conditions, including those involved in longitudinal disease progression.

To address this limitation, some recent works modify the contrastive framework to instead maximize the distance between the sampled clips \cite{dave2022tclr,zhuang2020unsupervised}, or by parameterizing the video with an evolving stochastic process \cite{park2022probabilistic,zhang2023modeling}. Yang et al. in Latent time navigation (LTN) \cite{yang2023self} introduce an extra orthogonal basis that explicitly models the temporal shift between sampled clips of the same video. Generative approaches, such as VideoMAE \cite{tong2022videomae} and v-JEPA \cite{bardes2023v}, implicitly model temporal variations by masking and reconstructing a subset of frame-level features. Still, it is unclear whether the ability of these methods to model temporal developments on natural videos translates to spatiotemporal medical data.

\subsection{Representation Learning for Spatiotemporal Medical Data}
In the medical field, spatiotemporal representation learning has many applications, such as organ segmentation \cite{shin2012stacked}, physiological development analysis \cite{zhao2021longitudinal}, and disease progression prediction \cite{couronne2021longitudinal}. Some works leverage the assumption that certain disease progressions share similar trajectories in the latent space \cite{ouyang2022self,ouyang2021self}. In addition, \cite{ren2022local} and \cite{wei2021consistent} leverage redundancies between successive brain scans to improve segmentation accuracy. However, there is no unified approach that learns self-supervised representations of spatiotemporal data in the medical domain. In this work, we propose a method that combines the efficacy of contrastive approaches with a masked latent reconstruction loss to model trajectories of frame-level temporal changes that are crucial for downstream medical analysis.
\section{Materials and Methods}

\subsection{Short and Long Time Scale Datasets}
\label{sec:dataset}
\subsubsection{Cardiac MR video}
To study temporal variation over short time frames we chose to use short-axis cardiac MR videos sourced from the UK Biobank (UKBB) \cite{sudlow2015uk} (see Figure \ref{fig:data}). 
The dataset contains 46,137 videos where each captures at least one full contraction, from end diastole to end systole, in 50 frames taken over 1.6 seconds.
After resampling the images to 1.8 mm in the sagittal and coronal dimensions, we used segmentation masks to crop a region of $128\times128$ about the center of mass of the hearts.

\subsubsection{Longitudinal Retinal OCT}
To study temporal change over long time frames using sparse, irregular data, we use an in-house retinal OCT dataset tracking patients with age-related macular degeneration (see Figure \ref{fig:data}) {collected by the Southampton Eye Unit in the scope of the PINNACLE study \cite{sutton2022developing}}. The dataset includes 48,825 OCT scans, monitoring 6,368 eyes in 3,498 patients. On average, each eye was scanned 7.7 times over 1.9 years, where eyes with AMD were observed for an average of 2.3 years.

\begin{figure}[tp]
    \centering
    \includegraphics[width=\linewidth]{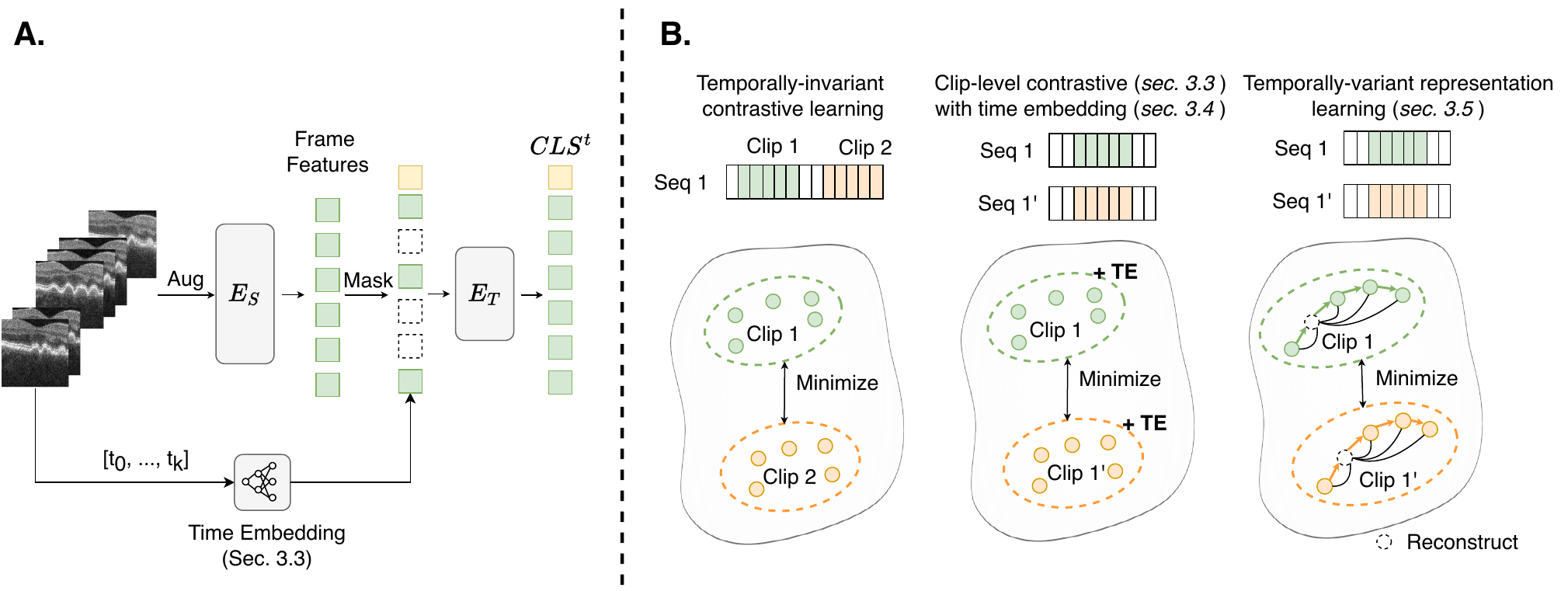}
    \caption{\textbf{A.} Our spatiotemporal encoder extracts representation of the sequence using the CLS token of the temporal Transformer $E_T$. 
    \textbf{B.} Standard contrastive approaches prioritize learning features that persist across different segments of the same sequence (left). Our approaches use a single clip to construct contrastive pairs combined with time embedding (middle) and a frame-level feature predictive approach (right) to model temporal variation over the sequence.}
    \label{fig:method}
\end{figure}

\subsection{Spatiotemporal Encoder}
As the basis of all baselines and our new approaches, we use a spatiotemporal encoder that extracts representation of the sequence (see Figure \ref{fig:method} A). For each image $x_i$ in the sequence, we employ a spatial Vision Transformer (ViT) \cite{dosovitskiy2020image} $E_S$ to extract a feature vector per image $z_i^s=E_S(x_i)$ using a learnable CLS token. The temporal Transformer $E_T$ then models temporal change of features and generates a global representation of the sequence $z^t = E_T(z^s_1 \dots z^s_i)$ using a distinct CLS token. Our two-encoder design efficiently operates on spatiotemporal data by reducing attention token length instead of processing all video patches at once.

\subsection{Clip-level Contrastive Learning}
\label{sec:contrastive learning}
In Figure \ref{fig:method} B., we illustrates that established contrastive learning methods prioritize temporally-invariant features by minimizing the feature distance between \textit{multiple} temporally distant clips from the same sequence \cite{qian2021spatiotemporal,feichtenhofer2021large}. To model temporal variations rather than temporal constants, we construct a clip-level contrastive approach that augments only a \textit{single} randomly sampled clip $c$. We then generate a pair of augmented views of our clip, $c_i$ and $c_j$, where the augmentations in each are applied consistently across timepoints. We pass these views through the spatiotemporal encoder to extract two clip-level contrastive tokens $z^t_i$ and $z^t_j$, which are learnable CLS tokens in the temporal encoder $E_T$. Finally, we optimize the NT-Xent \cite{chen2020simple} loss over the $N$ samples in the batch using equation \ref{eq:contrastive}, where $sim$ is the cosine similarity, $\tau$ is the temperature, $p$ is a non-linear projection head following the design of \cite{chen2020simple}.

\begin{equation}
L^C = -\log\frac{exp(sim(p(z^t_i), p(z^t_j))/{\tau})}{\sum_{k=1, k \neq i}^{2N} exp(sim(p(z^t_i), p(z^t_k))/{\tau})}
\label{eq:contrastive}
\end{equation}

\subsection{Time Embedding}
\label{sec:timeembedding}
To model the irregular time intervals between scans, we propose time embedding (TE) for the temporal Transformer $E_T$ (see Figure \ref{fig:method} B). For each clip, we calculate the relative time between each scan and the first scan. We then apply a learnable non-linear mapping based on Ho et al. \cite{ho2020denoising} to map the time points to the feature dimension of the model. Similar to a position embedding, we add TE to the input frame-level features in $E_T$. This enables the temporal encoder to estimate the rate of change in longitudinal sequences with irregular sampling.

\subsection{Temporally-Variant Feature Prediction}
\label{sec:novelty}
Finally, to explicitly model frame-level temporal variations, we propose a temporal feature prediction task inspired by Assran et al. \cite{assran2023self} (see Figure \ref{fig:method} B). Specifically, given a sequence of frame-level tokens output by our spatial encoder, we replace a random subset $Y$ with a learnable mask token using a masking ratio of 0.15. We then use a projection head $q$ on top of the temporal encoder output tokens to reconstruct the masked tokens. We then compute the cosine similarity loss using equation $\ref{eq:mlm}$ between the reconstructed tokens $Y^{\prime}$ and the original spatial tokens $Y$, where $m$ denotes the number of masked tokens.
\begin{equation}
\label{eq:mlm}
L^M = \frac{1}{m} \sum_{i=1}^m sim(Y^{\prime}_i, Y_i)
\end{equation}
The final loss, termed temporally-variant representation loss (TVRL), combines clip-level contrastive learning with frame-level feature prediction and is defined in equation \ref{eq:tvrl}, where $\lambda$ is a weight term empirically set to 0.5.
\begin{align}
\label{eq:tvrl}
L &= (1 - \lambda) L^C + \lambda L^M
\end{align}
\newcolumntype{?}{!{\vrule width 1.2pt}}
\section{Experiments and Results}
\subsection{Experimental Setup}
\subsubsection{Pretraining Protocol}
We evaluate our two pretraining approaches: the clip-level contrastive strategy described in section \ref{sec:contrastive learning} which is denoted as \textit{c}SimCLR, and our TVRL extension introduced in section \ref{sec:novelty}. For \textit{c}SimCLR, we also evaluate the effect of adding time embedding (TE) from section \ref{sec:timeembedding} and denote it as \textit{c}SimCLR-TE. Note that our TVRL strategy does not use TE. We leverage standard SimCLR augmentations \cite{chen2020simple} and discard those unsuitable for grayscale images. We compare our approaches against the leading spatiotemporal representation learning strategies, which includes CVRL \cite{qian2021spatiotemporal}, LTN \cite{yang2023self}, and VideoMAE \cite{tong2022videomae}. All pretraining strategies use a spatial encoder $E_S$ parameterized by a ViT-S \cite{dosovitskiy2020image} with 384 hidden dimensions, six heads, and 12 attention layers, while the temporal encoder $E_T$ shares the same architecture but with three layers of attention. Both encoders are randomly initialized before pretraining. In total the model has 27.7M parameters.

Before pretraining, we first partition each dataset into training, validation, and test sets, comprising 70\%, 15\%, and 15\% of the data, respectively, ensuring that each patient appears in only one subset. Each model accepts eight images as input. We sample every other frame from the cardiac videos, and every successive scan in the retinal longitudes. We employ separate projection heads $p$ and $q$ for our contrastive and masked reconstruction losses, following the two-layer configuration in \cite{chen2020simple}. All models are pretrained for 200 epochs with a batch size of 256 using the AdamW optimizer, and a cosine annealing learning rate of $2 \cdot 10^{-4}$ with 20 epochs of linear warm-up.

\subsubsection{Finetuning Protocol}
We follow the standard linear probing protocol \cite{DBLP:journals/corr/abs-2002-05709} and finetune each pretrained model by freezing the network and appending a new trainable CLS token to the temporal Transformer which serves as input to a single linear layer. We finetune each model with 1\% and 100\% of the data, and report the area under the receiver operating characteristic curve (AUC) for classification tasks and mean absolute error (MAE) for regression tasks, with mean and standard deviation over five random seeds. We finetune with AdamW for 100 epochs with a learning rate of $10^{-3}$ and a batch size of 256 and report test performance of the best validation epoch. As baselines to the aforementioned strategies, we also evaluate a fully supervised spatiotemporal encoder trained end-to-end, in addition to a 3D ResNet50 \cite{feichtenhofer2019slowfast} (31.6M parameters) pre-trained on the Kinetics-400 dataset \cite{kay2017kinetics}.
During inference, in sequences with more than eight images, we adapt the common practice in evaluating video models \cite{feichtenhofer2021large,qian2021spatiotemporal} and apply a sliding window with 50\% overlap to contiguous clips before averaging these predictions. 

\subsection{Downstream Evaluation}
We first evaluate different pretraining strategies on a set of control tasks that are marginally associated with temporal change. In cardiac MR video, we predict coronary artery disease (CAD) and fibrillation from ICD codes recorded during hospital admissions of all subjects. From retinal OCT longitudinal data we predict the presence of late vs. early AMD given the present and historical set of scans of each patient.

We then evaluate the models on tasks that are strongly associated with temporal change. In the short time scale, we estimate left ventricular ejection fraction (LVEF) and total cardiac output, which measures the volume of blood pumped in one minute. Finally, we focus on three OCT-based prognostic tasks that require extrapolating long term trends from the patient's history of disease progression. These involve predicting the development of late stage wet AMD (defined by choroidal neovascularisation), late stage dry AMD (defined by cRORA > 1000$\mu m$ \cite{sadda2018consensus}) and retinal scarring and fibrosis post treatment, all within the next 3 years.

\begin{table}[t]
\centering
\caption{Performance on diagnosis and prognosis on the long term AMD prediction tasks in longitudinal retinal OCT. Modeling variable intervals between scans is especially beneficial for prognosis tasks that requires the extrapolation of trajectories of disease progression captured in historical scans.}
\label{tab:oct}

\resizebox{\textwidth}{!}{%
\begin{tabular}{c?cc?cc|cc|cc}
\hline
 & \multicolumn{2}{c?}{\begin{tabular}[c]{@{}c@{}} \textbf{Non-temporal} \\ \textbf{control task}\vspace{0.1in} \end{tabular}}  & 
 \multicolumn{6}{c}{\begin{tabular}[c]{@{}c@{}} \textbf{Temporally dependent prognostic tasks} \vspace{0.1in}\end{tabular}}  \\

Model &  \multicolumn{2}{c?}{\begin{tabular}[c]{@{}c@{}}Late vs. Early \\ AUC (\%) $\uparrow$ \end{tabular}} & \multicolumn{2}{c|}{\begin{tabular}[c]{@{}c@{}}Late Dry AMD\\ AUC (\%) $\uparrow$ \end{tabular}} & \multicolumn{2}{c|}{\begin{tabular}[c]{@{}c@{}}Late Wet AMD\\ AUC (\%) $\uparrow$ \end{tabular}} & \multicolumn{2}{c}{\begin{tabular}[c]{@{}c@{}}Scarring and Fibrosis\\ AUC (\%) $\uparrow$ \end{tabular}}\\
 & 1\% & 100\% & 1\% & 100\% & 1\% & \multicolumn{1}{c|}{100\%} & 1\% & 100\%  \\ \hline \hline

Supervised & $60.2_{\pm 1.1}$ & $85.9_{\pm 0.3}$ & $50.2_{\pm 1.1}$ & $64.7_{\pm 0.5}$ & $54.5_{\pm 0.2}$ & $59.2_{\pm 0.9}$ & $49.1_{\pm 0.8}$ & $52.9_{\pm 0.6}$ \\
R3D \cite{feichtenhofer2019slowfast} (Kinetics)  & $54.3_{\pm 1.1}$ & $73.8_{\pm 0.3}$ & $51.8_{\pm 5.1}$ & $54.4_{\pm 0.6}$ & $52.2_{\pm 2.1}$ & $64.9_{\pm 0.8}$ & $50.6_{\pm 5.1}$ & $64.4_{\pm 1.3}$ \\
CVRL \cite{qian2021spatiotemporal} & $75.3_{\pm 1.7}$ & $86.2_{\pm 0.3}$ & $52.4_{\pm 2.3}$ & $67.5_{\pm 5.3}$ & $56.8_{\pm 6.5}$ & $64.7_{\pm 2.6}$ & $59.9_{\pm 3.9}$ & $71.2_{\pm 1.7}$ \\
LTN \cite{yang2023self}& $\mathbf{76.9_{\pm 0.6}}$ & $85.7_{\pm 0.1}$ & $\mathbf{61.8_{\pm 3.4}}$ & $63.1_{\pm 1.5}$ & $57.6_{\pm 6.1}$ & $64.8_{\pm 4.1}$ & $\mathbf{71.6_{\pm 0.9}}$ & $72.9_{\pm 0.8}$ \\
VideoMAE \cite{tong2022videomae}  & $72.0_{\pm 1.1}$ & $81.8_{\pm 0.1}$ & $\underline{58.7_{\pm 4.1}}$ & $56.8_{\pm 0.6}$ & $54.3_{\pm 9.2}$ & $56.2_{\pm 7.8}$ & $61.4_{\pm 3.4}$ & $67.3_{\pm 1.1}$ \\ \hline

\textit{c}SimCLR (ours)  & $\underline{76.3_{\pm 2.7}}$ & $\mathbf{88.6_{\pm 0.4}}$ & $53.6_{\pm 2.8}$ & $74.3_{\pm 1.3}$ & $\underline{58.3_{\pm 4.2}}$ & $\underline{66.8_{\pm 1.3}}$ & $\underline{68.9_{\pm 4.5}}$ & $72.2_{\pm 0.6}$ \\
\textit{c}SimCLR-TE (ours)  & $73.0_{\pm 0.9}$ & ${\underline{88.3_{\pm 0.1}}}$ & $50.4_{\pm 5.6}$ & $\underline{76.8_{\pm 1.5}}$ & $\mathbf{58.7_{\pm 5.5}}$ & {${66.5_{\pm 1.1}}$} & $64.4_{\pm 3.3}$ & $\mathbf{77.4_{\pm 1.5}}$ \\
TVRL (ours) & $76.1_{\pm 1.6}$ & $87.7_{\pm 0.1}$ & $55.7_{\pm 4.0}$ & $\mathbf{80.2_{\pm 0.8}}$ & $53.5_{\pm 4.8}$ & $\mathbf{67.8_{\pm 2.3}}$ & $65.9_{\pm 3.0}$ & $\underline{73.2_{\pm 1.4}}$ \\

\hline
\end{tabular}
}
\end{table}

\begin{table}[h!]
\centering
\caption{In short term cardiac video, standard natural video approaches perform comparably on diagnosis but underperform in estimating cardiac output, which has the strongest temporal component.}
\label{tab:cardiac}
\resizebox{\textwidth}{!}{
\begin{tabular}{c?cc|cc?cc|cc}
\hline
 & \multicolumn{4}{c?}{\begin{tabular}[c]{@{}c@{}} \textbf{Diagnostic control tasks} \vspace{0.1in}\end{tabular}}  & 
 \multicolumn{4}{c}{\begin{tabular}[c]{@{}c@{}} \textbf{Temporally dependent biometric tasks} \vspace{0.1in}\end{tabular}} \\

Model & \multicolumn{2}{c|}{\begin{tabular}[c]{@{}c@{}}CAD\\  AUC (\%) $\uparrow$ \end{tabular}} & \multicolumn{2}{c?}{\begin{tabular}[c]{@{}c@{}}Fibrillation \\ AUC (\%) $\uparrow$ \end{tabular}} & \multicolumn{2}{c|}{\begin{tabular}[c]{@{}c@{}}LVEF \\ MAE $\downarrow$ \end{tabular}} & \multicolumn{2}{c}{\begin{tabular}[c]{@{}c@{}}Cardiac Output \\ MAE $\downarrow$ \end{tabular}} \\ 
 & 1\% & 100\% & 1\% & 100\% & 1\% & 100\% & 1\% & 100\% \\ \hline \hline
Supervised & $56.3_{\pm 0.8}$ & $64.9_{\pm 3.1}$ & $53.2_{\pm 7.2}$ & $62.7_{\pm 0.6}$ & $4.75_{\pm 0.01}$ & $3.58_{\pm 0.21}$ & $4.33_{\pm 3.38}$ & $1.80_{\pm 0.04}$\\
R3D \cite{feichtenhofer2019slowfast} (Kinetics) & $52.9_{\pm 1.8}$ & $64.1_{\pm 4.2}$ & $54.6_{\pm 0.2}$ & $66.3_{\pm 2.1}$ & $8.27_{\pm 1.24}$ & $7.13_{\pm 0.58}$ & $2.77_{\pm 0.09}$ & $1.86_{\pm 0.05}$ \\ 
CVRL \cite{qian2021spatiotemporal} & $59.4_{\pm 5.8}$ & $\underline{69.4_{\pm 0.2}}$ & $54.4_{\pm 0.9}$ & $70.3_{\pm 0.3}$ & $5.14_{\pm 0.24}$ & $3.95_{\pm 0.03}$ & \underline{$1.85_{\pm 0.02}$} & $1.76_{\pm 0.08}$ \\
LTN \cite{yang2023self} & $56.6_{\pm 8.7}$ & $\underline{69.4_{\pm 0.2}}$ & $54.0_{\pm 1.3}$ & $\mathbf{70.5_{\pm 0.3}}$ & $\underline{4.83_{\pm 0.04}}$ & $\mathbf{3.88_{\pm 0.08}}$ & \underline{$1.85_{\pm 0.01}$} & $1.68_{\pm 0.03}$ \\
VideoMAE \cite{tong2022videomae} & $57.1_{\pm 4.3}$ & $66.9_{\pm 0.6}$ & $55.1_{\pm 3.1}$ & $63.1_{\pm 0.9}$ & $6.89_{\pm 0.05}$ & $5.28_{\pm 0.07}$ & $1.98_{\pm 0.02}$ & $1.86_{\pm 0.06}$ \\ \hline

\textit{c}SimCLR (ours) & $60.9_{\pm 0.4}$ & $\underline{69.4_{\pm 0.1}}$ & $54.5_{\pm 0.4}$ & $\underline{70.4_{\pm 0.6}}$ & $4.88_{\pm 0.35}$ & $\underline{3.89_{\pm 0.04}}$ & $1.88_{\pm 0.01}$ & $1.69_{\pm 0.12}$ \\
\textit{c}SimCLR-TE (ours) & $\mathbf{62.6_{\pm 0.9}}$ & $67.9_{\pm 0.5}$ & $\mathbf{57.3_{\pm 0.2}}$ & $66.7_{\pm 0.1}$ & $\mathbf{4.62_{\pm 0.01}}$ & $4.05_{\pm 0.01}$ & $\mathbf{1.84_{\pm 0.01}}$ & $\mathbf{1.63_{\pm 0.01}}$ \\
TVRL (ours) & $\underline{61.7_{\pm 0.1}}$ & $69.3_{\pm 0.1}$ & \underline{$55.5_{\pm 0.7}$} & $69.8_{\pm 0.1}$ & $4.86_{\pm 0.02}$ & $4.03_{\pm 0.01}$ & $1.92_{\pm 0.02}$ & \underline{$1.65_{\pm 0.02}$} \\
\hline
\end{tabular}
}
\end{table}

\subsection{Simple Clip-level Contrastive Learning Outperforms Prior Approaches}
\label{subsec:contrastive repre}
In Table \ref{tab:oct} and Table \ref{tab:cardiac}, we find in all tasks that the time-invariant representations learned by CVRL \cite{qian2021spatiotemporal} degrade performance compared to our simple time-dependent \textit{c}SimCLR \cite{chen2020simple} approach using a single clip. 
This was most pronounced on the temporally dependent tasks such as prognosis of late dry AMD (74.3\% vs. 67.5\% AUC). 
While LTN \cite{yang2023self} and VideoMAE \cite{tong2022videomae} improve the modeling of temporal variations, both methods perform comparably to or worse than clip-level \textit{c}SimCLR across all tasks in both the short and long time scales. 

\subsection{Explicit Temporal Modeling Improves Long-term Prognostic Tasks}
The clip-level contrastive approach with time embedding (\textit{c}SimCLR-TE) outperforms all prior methods on the irregularly sampled retinal OCT longitudes. Add TE notably improves the prognosis of scarring from 72.2\% to 77.4\% AUC as shown in Table \ref{tab:oct}. It also performed best in estimating the cardiac output, achieving 1.63 MAE as shown in Table \ref{tab:cardiac}. Finally, TVRL further boosts the performance on prognosis for late dry AMD to 80.2\% AUC and late wet AMD to 67.8\% AUC. Additionally, in Figure A.1 and A.2 in the supplementary, we visualize the feature trajectories of sequences from both datasets learned by TVRL.

Regarding Table \ref{tab:cardiac}, on cardiac videos with regular frame intervals, TE is expected to provide no further benefit over positional embedding, leading to comparable performance between \textit{c}SimCLR and \textit{c}SimCLR-TE. Similarly, TVRL performs comparably to contrastive baselines, as cardiac videos contain limited temporal variation within a single cardiac cycle.
\section{Discussion and Conclusion}
In this paper, we identified two common variants of spatiotemporal data used in clinical practice. We found that while state-of-the-art spatiotemporal learning methods developed for natural videos perform comparably in short term analysis on cardiac MR video, they failed to track long term developments and disease trajectories in longitudinal series, leading to poor prognostic utility for AMD. To address this, we propose a simple clip-level contrastive learning strategy that leverages time embeddings in irregular and variable length time series, and a new temporally-variant approach that explicitly models frame-level variation. Models pretrained with our strategies had improved assessment of cardiac output, and substantially improved prognosis for three variants of late stage AMD.

One limitation of this study is that we have not fully determined in which scenario to use each of our two approaches. Future work aims to expand the number and diversity of tasks to provide better guidance.

In conclusion, we envision that our approaches will create representations that leverage the full range of temporal dynamics and patient history that are typically neglected in current diagnostic and prognostic solutions.
\begin{credits}
\subsubsection{Acknowledgments}
This work is funded by a Wellcome Trust Collaborative Award (ref. 210572/Z/18/Z), and an EPSRC grant (ref. EP/Y015665/1). The authors acknowledge support from the German Research Foundation (project 532139938) and the European Research Council (Deep4MI, Grant Agreement no. 884622). This research has been conducted using the UK Biobank dataset under the application number 87802.
\end{credits}

\begin{credits}
\subsubsection{\discintname}
The authors have no competing interests to declare that are
relevant to the content of this article.
\end{credits}

\newpage
\bibliographystyle{splncs04}
\bibliography{bibliography}

\begin{thebibliography}{10}
\providecommand{\url}[1]{\texttt{#1}}
\providecommand{\urlprefix}{URL }
\providecommand{\doi}[1]{https://doi.org/#1}

\bibitem{assran2023self}
Assran, M., et~al.: Self-supervised learning from images with a joint-embedding predictive architecture. In: CVPR. pp. 15619--15629 (2023)

\bibitem{bardes2023v}
Bardes, A., et~al.: V-jepa: Latent video prediction for visual representation learning  (2023)

\bibitem{bernard2018deep}
Bernard, O., et~al.: Deep learning techniques for automatic mri cardiac multi-structures segmentation and diagnosis: is the problem solved? IEEE TMI  \textbf{37}(11),  2514--2525 (2018)

\bibitem{caron2021emerging}
Caron, M., et~al.: Emerging properties in self-supervised vision transformers. In: CVPR. pp. 9650--9660 (2021)

\bibitem{chainani2001oral}
Chainani-Wu, N., et~al.: Oral lichen planus: patient profile, disease progression and treatment responses. The Journal of the American Dental Association  \textbf{132}(7),  901--909 (2001)

\bibitem{DBLP:journals/corr/abs-2002-05709}
Chen, T., Kornblith, S., Norouzi, M., Hinton, G.E.: A simple framework for contrastive learning of visual representations. CoRR  \textbf{abs/2002.05709} (2020), \url{https://arxiv.org/abs/2002.05709}

\bibitem{chen2020simple}
Chen, T., et~al.: A simple framework for contrastive learning of visual representations. In: ICML. pp. 1597--1607. PMLR (2020)

\bibitem{chen2021exploring}
Chen, X., et~al.: Exploring simple siamese representation learning. In: CVPR. pp. 15750--15758 (2021)

\bibitem{couronne2021longitudinal}
Couronn{\'e}, R., et~al.: Longitudinal self-supervision to disentangle inter-patient variability from disease progression. In: MICCAI. pp. 231--241. Springer (2021)

\bibitem{dave2022tclr}
Dave, I., et~al.: Tclr: Temporal contrastive learning for video representation. CVIU  \textbf{219},  103406 (2022)

\bibitem{dosovitskiy2020image}
Dosovitskiy, A., et~al.: An image is worth 16x16 words: Transformers for image recognition at scale. arXiv preprint arXiv:2010.11929  (2020)

\bibitem{feichtenhofer2019slowfast}
Feichtenhofer, C., Fan, H., Malik, J., He, K.: Slowfast networks for video recognition. In: ICCV. pp. 6202--6211 (2019)

\bibitem{feichtenhofer2021large}
Feichtenhofer, C., et~al.: A large-scale study on unsupervised spatiotemporal representation learning. In: CVPR. pp. 3299--3309 (2021)

\bibitem{graham2014current}
Graham, L.J., et~al.: Current approaches and challenges in monitoring treatment responses in breast cancer. Journal of Cancer  \textbf{5}(1), ~58 (2014)

\bibitem{grill2020bootstrap}
Grill, J.B., et~al.: Bootstrap your own latent-a new approach to self-supervised learning. NeurIPS  \textbf{33},  21271--21284 (2020)

\bibitem{he2020momentum}
He, K., et~al.: Momentum contrast for unsupervised visual representation learning. In: CVPR. pp. 9729--9738 (2020)

\bibitem{ho2020denoising}
Ho, J., Jain, A., Abbeel, P.: Denoising diffusion probabilistic models. Advances in neural information processing systems  \textbf{33},  6840--6851 (2020)

\bibitem{holland2023clustering}
Holland, R., et~al.: Clustering disease trajectories in contrastive feature space for biomarker proposal in age-related macular degeneration. In: MICCAI. pp. 724--734. Springer (2023)

\bibitem{jensen2014temporal}
Jensen, A.B., et~al.: Temporal disease trajectories condensed from population-wide registry data covering 6.2 million patients. Nature communications  \textbf{5}(1), ~4022 (2014)

\bibitem{kay2017kinetics}
Kay, W., et~al.: The kinetics human action video dataset. arXiv preprint arXiv:1705.06950  (2017)

\bibitem{koch2013fear}
Koch, L., et~al.: Fear of recurrence and disease progression in long-term ($\geq$ 5 years) cancer survivors—a systematic review of quantitative studies. Psycho-oncology  \textbf{22}(1),  1--11 (2013)

\bibitem{komatsu2021detection}
Komatsu, M., et~al.: Detection of cardiac structural abnormalities in fetal ultrasound videos using deep learning. Applied Sciences  \textbf{11}(1), ~371 (2021)

\bibitem{mondejar2019heartbeat}
Mond{\'e}jar-Guerra, V., et~al.: Heartbeat classification fusing temporal and morphological information of ecgs via ensemble of classifiers. Biomedical Signal Processing and Control  \textbf{47},  41--48 (2019)

\bibitem{ouyang2021self}
Ouyang, J., et~al.: Self-supervised longitudinal neighbourhood embedding. In: MICCAI. pp. 80--89. Springer (2021)

\bibitem{ouyang2022self}
Ouyang, J., et~al.: Self-supervised learning of neighborhood embedding for longitudinal mri. Medical image analysis  \textbf{82},  102571 (2022)

\bibitem{park2022probabilistic}
Park, J., et~al.: Probabilistic representations for video contrastive learning. In: CVPR. pp. 14711--14721 (2022)

\bibitem{pugash2008prenatal}
Pugash, D., et~al.: Prenatal ultrasound and fetal mri: the comparative value of each modality in prenatal diagnosis. European journal of radiology  \textbf{68}(2),  214--226 (2008)

\bibitem{qian2021spatiotemporal}
Qian, R., et~al.: Spatiotemporal contrastive video representation learning. In: CVPR. pp. 6964--6974 (2021)

\bibitem{ren2022local}
Ren, M., et~al.: Local spatiotemporal representation learning for longitudinally-consistent neuroimage analysis. NeurIPS  \textbf{35},  13541--13556 (2022)

\bibitem{sadda2018consensus}
Sadda, S.R., et~al.: Consensus definition for atrophy associated with age-related macular degeneration on oct: classification of atrophy report 3. Ophthalmology  \textbf{125}(4),  537--548 (2018)

\bibitem{shin2012stacked}
Shin, H.C., et~al.: Stacked autoencoders for unsupervised feature learning and multiple organ detection in a pilot study using 4d patient data. IEEE TPAMI  \textbf{35}(8),  1930--1943 (2012)

\bibitem{sudlow2015uk}
Sudlow, C., et~al.: Uk biobank: an open access resource for identifying the causes of a wide range of complex diseases of middle and old age. PLoS medicine  \textbf{12}(3),  e1001779 (2015)

\bibitem{sutton2022developing}
Sutton, J., et~al.: Developing and validating a multivariable prediction model which predicts progression of intermediate to late age-related macular degeneration—the pinnacle trial protocol. Eye pp.~1--9 (2022)

\bibitem{tong2022videomae}
Tong, Z., et~al.: Videomae: Masked autoencoders are data-efficient learners for self-supervised video pre-training. NeurIPS  \textbf{35},  10078--10093 (2022)

\bibitem{wang2022long}
Wang, J., et~al.: Long-short temporal contrastive learning of video transformers. In: CVPR. pp. 14010--14020 (2022)

\bibitem{wei2021consistent}
Wei, J., et~al.: Consistent segmentation of longitudinal brain mr images with spatio-temporal constrained networks. In: MICCAI. pp. 89--98. Springer (2021)

\bibitem{yang2023self}
Yang, D., et~al.: Self-supervised video representation learning via latent time navigation. arXiv preprint arXiv:2305.06437  (2023)

\bibitem{zhang2023modeling}
Zhang, H., et~al.: Modeling video as stochastic processes for fine-grained video representation learning. In: CVPR. pp. 2225--2234 (2023)

\bibitem{zhao2021longitudinal}
Zhao, Q., et~al.: Longitudinal correlation analysis for decoding multi-modal brain development. In: MICCAI. pp. 400--409. Springer (2021)

\bibitem{zhuang2020unsupervised}
Zhuang, C., et~al.: Unsupervised learning from video with deep neural embeddings. In: CVPR. pp. 9563--9572 (2020)

\end{thebibliography}

\end{document}


\renewcommand{\thefigure}{A.\arabic{figure}}
\setcounter{figure}{0}

\section*{Supplementary Material}

\begin{figure}[h!]
    \centering
    \includegraphics[width=\linewidth]{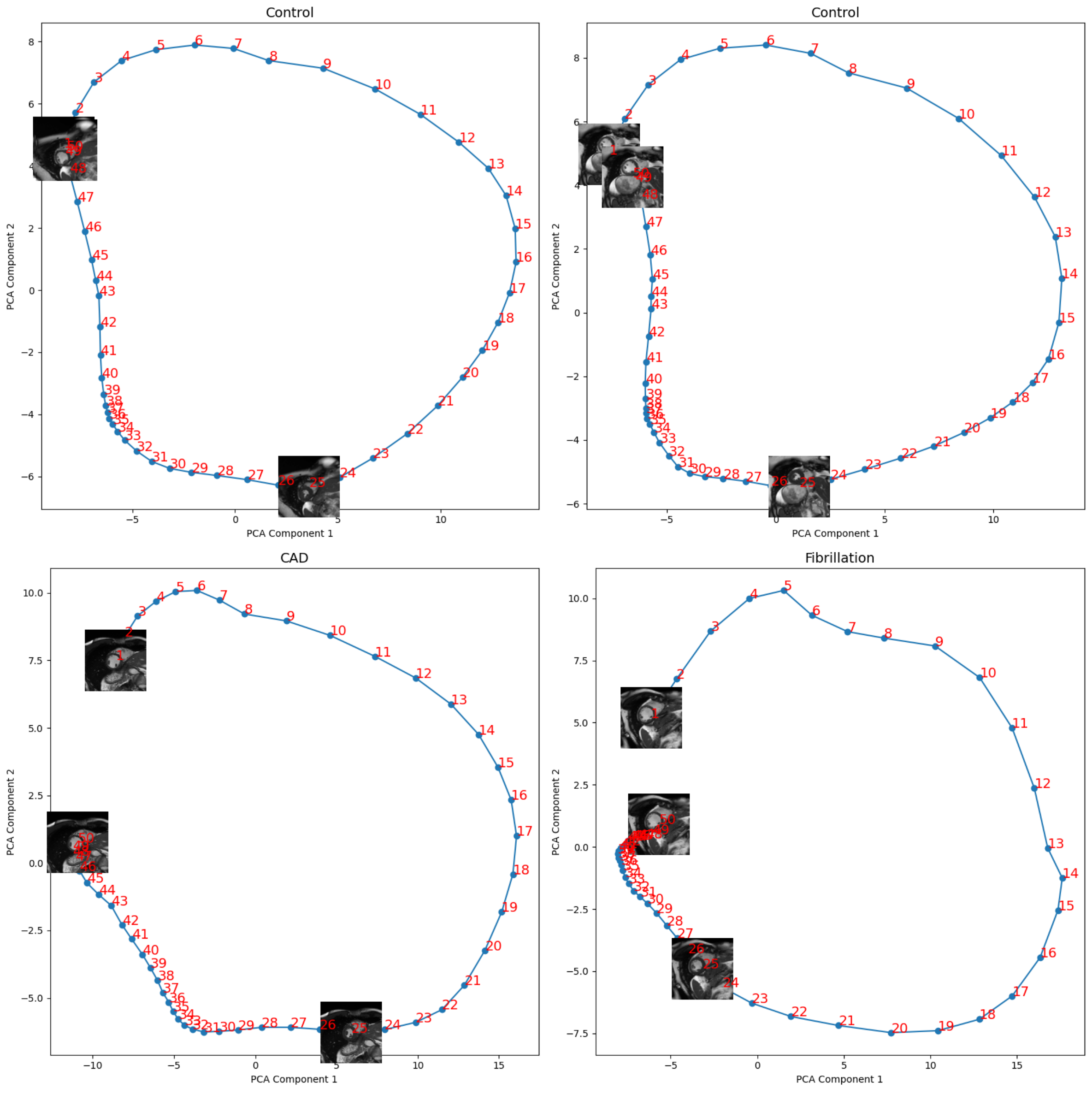}
    \caption{Visualization of the trajectory of cardiac videos with two control samples and two types of diseases. We plot the first two PCA components of the latent space of the temporal encoder $E_T$ pretrained with our TVRL strategy. Each point on the trajectory represents a frame in the video annotated with the frame index. \textbf{Top row:} The trajectories in control samples form a circular shape, matching a periodic pattern in the cardiac cycle. \textbf{Bottom row:} There is a notable gap for the embeddings of the sample with CAD between its first and last frame, while the embeddings of fibrillation are irregularly distributed due to irregular heart rhythms. Additionally, we visualize the first, middle, and last frames of each video to provide more visual context.}
    \label{fig:cardiac_trajectory}
\end{figure}

\begin{figure}[h!]
    \centering
    \includegraphics[width=\linewidth]{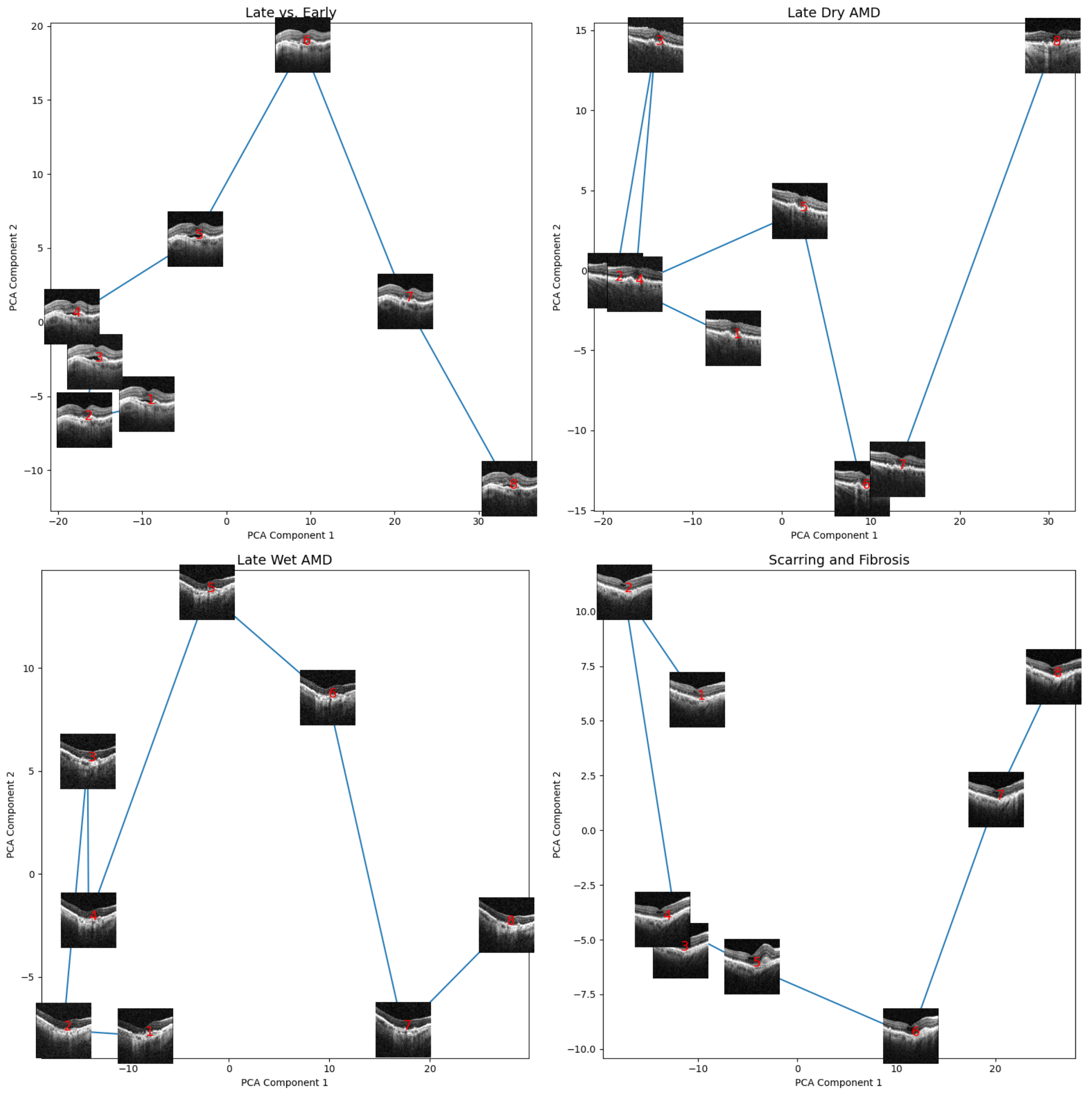}
    \caption{   
    Visualization of the trajectory of longitudinal retinal OCT scans with four types of diseases. We plot the first two PCA components of the latent space of the temporal encoder $E_T$ pretrained with our TVRL strategy. Each point on the trajectory corresponds to a scan in the longitudinal sequence. The AMD disease progression has different conversion rates, typically with an initial health stage deterioration (e.g. images 4 and 5 in the plot for scarring at bottom right) before disease conversion.
  }
    \label{fig:oct_trajectory}
\end{figure}